# Does Thought Require Sensory Grounding? From Pure Thinkers to Large Language Models

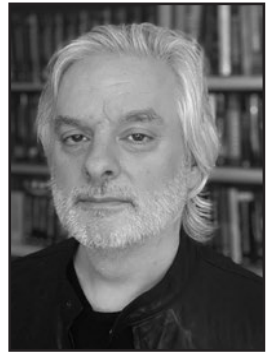

David J. Chalmers
**NEW YORK UNIVERSITY**

Does the capacity to think require the capacity to sense? A lively debate on this topic runs throughout the history of philosophy and now animates discussions of artificial intelligence.

In favor of a positive answer, Aristotle says, "The soul never thinks without an image." Aquinas says, "There's nothing in the intellect that wasn't previously in the senses." Hume says, "All our simple ideas in their first appearance are derived from simple impressions." With some minimal assumptions, all three of these statements suggest that thinking requires the capacity to sense, or at least requires having had the capacity to sense at some point.

Contrasting with these empiricist theses, rationalist philosophers have often denied that thinking requires sensing. Plato holds that we can think about the forms before we have senses and a body. Descartes holds that the pure intellect thinks independently of the senses. Navigating between empiricism and rationalism, Kant discusses the issue extensively ("Thoughts without content are empty"); unsurprisingly, his final views on the matter are complicated.

In recent decades, this philosophical debate has become central to debates in artificial intelligence and cognitive science. The cognitive scientist Stevan Harnad (1990) put forward the symbol grounding problem: How do symbols in AI systems come to mean anything? He and others held that for symbols to have meaning, they must be causally grounded in sensory connections to the environment. To be meaningful, the symbol "RED" must be grounded in seeing red. The symbol "WATER" must be grounded in a sensory connection to water. If we assume that thinking and meaning go together in AI systems, then this amounts to another version of the thesis that thinking requires sensing.

In the last few years, discussion of symbol grounding has become especially widespread in the debate over large language models (LLMs) such as the GPT systems. Can large language models think, mean, or understand? Many researchers argue that they cannot, precisely because their symbols lack appropriate grounding.

In their well-known critique of meaning and understanding in language models, the computational linguists Emily Bender and Alexander Koller (2020) have argued that "a system that is trained only on form [such as an LLM] would fail a sufficiently sensitive test [for intelligence], because it lacks the ability to connect its utterances to the world."[1] Likewise, the psychologists Brenden Lake and Gregory Murphy (2023) have argued that "word meaning in both human and AI systems must be grounded in perception and action."[2] The philosopher Jacob Browning and the AI researcher Yann LeCun (2022) have argued that "LLMs have no stable body or abiding world to be sentient *of*—so their knowledge begins and ends with more words and their common-sense is skin deep."

These researchers make various different claims, but all appear to share the core view that since language models lack a grounding in the world, their ability to think (or mean or understand) is deeply limited if not absent altogether. We could put one version of the argument as follows:

> 1. Language models lack sensory capacities.
>
> 2. Genuine thought requires sensory capacities.
>
> ______________________
>
> So: 3. Language models lack genuine thought.

Here, the key premise is our original thesis that thinking requires sensing. Of course, a proponent of language models might react in many ways. They might reject premise 1 by arguing that LLMs have textual input systems which count as a sort of sense. They might reject premise 2 by arguing that LLMs don't need senses in order to think. They could accept the conclusion where pure language models are concerned, while arguing that multimodal language models, which process images and audio data and control a body, have sensory capacities and thereby avoid the argument. Still, this is an important critique of language models that requires analysis.

In what follows, I will argue against the thesis that thinking and understanding require sensing. Drawing on the history of philosophy, I will argue that in principle, there can be highly sophisticated thinkers that lack the capacity to sense altogether. That said, there are significant limitations in just what sort of thought is possible in the absence of the capacity to sense. I will explore these limitations, too. My case will be more of a brisk overview than a decisive argument, but I hope to at least set up a challenge for opponents to respond to. Toward the end, I will extend the analysis from thinking to understanding, and I will address the upshot for large language models and other AI systems. My aim is limited: I will not argue directly that large language models can think or understand. But I will at least rebut one important argument that they cannot.

## THE SENSE-THOUGHT THESIS

The primary thesis at issue is what I'll call the sense-thought thesis:

> Sense-Thought Thesis: Thinking requires having had the capacity to sense.

We could put the thesis a little more precisely by saying that necessarily, if S thinks a thought at time t, then S was able to sense at some time at or before t. The rest of this section clarifies various elements of the thesis (if you don't care about these details, feel free to skip them).

Regarding "having had" and "at or before t": there are stronger sense-thought theses saying that thinking requires concurrent sensing or at least a concurrent capacity to sense. But the past-directed "having had" thesis comes closer to capturing the formulations from Aquinas and Hume above, which seem to require that thinking be grounded in previous sensing. The past-directed thesis also plausibly allows that a being could lose the capacity to sense while continuing to think thoughts that were grounded in its prior sensory capacities. An alternative present-directed thesis without "having had," more in the spirit of Aristotle's formulation, says that all thinking requires (and perhaps is grounded in) concurrent sensory capacities or quasi-sensory capacities, where quasi-sensory capacities are modality-specific capacities such as visual imagery that typically derive from the senses but can survive the loss of strictly sensory capacities such as vision. The past-directed and present-directed sense-thought theses can come apart in some hard cases,[3] and each may be endorsed by different sensory grounding theorists. For my purposes the differences between them will not be central, and both theses are available.

What is the scope of the thesis? The *human* sense-thought thesis applies only to humans. The *unrestricted* thesis applies to all possible thinkers. The human thesis has been the subject of by far the most discussion to date, both in the history of philosophy and in cognitive psychology. I'll depart from these traditions by focusing mainly on the unrestricted thesis, in part because this thesis applies to AI systems while the human thesis does not. It's notable that while Aquinas and Aristotle accept the human sense-thought thesis, they reject the unrestricted thesis, since they hold that angels (not to mention God) can think without the capacity to sense. The fact that many empiricists and rationalists alike reject the unrestricted thesis perhaps gives that rejection some initial historical support.

I have put the sense-thought thesis in terms of possibility and necessity: Is it *possible* to think without the capacity to sense? Arguably, a more fundamental issue concerns grounding: Is thinking always grounded in the capacity to sense? For philosophers these days, the relevant sort of grounding is usually constitutive (though the historical quotes from Aquinas and Hume are not entirely clear on causal vs. constitutive grounding). For cognitive scientists, the relevant sort of grounding is sometimes constitutive and sometimes causal. For present purposes, it is more straightforward and less technical to cast things in terms of possibility rather than grounding. If it is even possible to think without having had the capacity to sense (or without having quasi-sensory capacities), as I will argue, it follows straightforwardly that thinking need not be grounded in sensory (or quasi-sensory) capacities. Still, considerations about grounding are never far away in my discussion.

It remains to clarify sensing and thinking. I take thinking to include mental acts such as judging and wondering, as well as dispositional mental states such as believing and desiring. I take it that thinking requires concepts, and that at least in the core cases I'll be concerned with, thinking is a propositional attitude: it involves an attitude (such as judging) to a proposition (e.g., that the sky is blue). In practice, I'll take judging as the paradigm case of thinking.

Sensing is more complicated. What counts as sensing, exactly? There are many strands in the notion. One strand is tied to *input*: senses provide inputs from the outside world. One strand is tied to *experience*: senses involve a certain sort of rich sensory experience. One strand is tied to *representation*: senses involve a certain sort of analog or iconic mental representation. Paradigm examples of sensing involve all three strands. On the other hand, there are cases where each is missing. Imagery need not involve inputs. Unconscious perception need not involve sensory experience. Digital sensory systems need not involve analog or iconic representation.

What sort of sensing do we require for the sense-thought thesis? Requiring all three factors (input, sensory experience, analog representation) seems too demanding. Different sense-thought proponents might stress different factors. Externalist symbol grounding proponents require especially environmental inputs. Image theorists may stress analog representation. Some phenomenal theorists may stress qualitative sensory experience. For a working definition, I will stipulate initially that any one of these three factors suffices for sensing. This inclusive definition of sensing yields a relatively weak version of the sense-thought thesis, which seems fairest for the purposes of arguing against the thesis. For certain purposes, I'll be more specific.

There are some further questions about what counts as sensing for our purposes of assessing the sense-thought thesis. Does introspection count as a sense? I'll start by assuming that it does not, but I'll also pay attention to how the issue changes if introspection is counted as a sense. Do bodily senses (e.g., proprioception and interoception) count as senses? Yes, absolutely. Including bodily senses allows proponents of the sense-thought thesis to include many bodily grounding theorists (proponents of the grounding of all thought in the body) as well as many action grounding theorists (proponents of the grounding of all thought in motor action). To include the latter, we could also cast the sense-thought thesis in terms of sensorimotor capacities as opposed to merely sensory capacities. My arguments will still go through just as well.

## PURE THINKERS

The sense-thought thesis turns on whether *pure thinkers* are possible. Pure thinkers, I'll stipulate, are beings that can think but that have never had the capacity to sense, and that lack even quasi-sensory capacities such as imagery. The name is reminiscent of Descartes's "pure intellects." Both labels might seem to exalt the beings, but I do not intend that rationalist implication. We could equally call pure thinkers *mere thinkers*. They are limited beings that can think while lacking sensory capacities altogether.

The classic case of something like a pure thinker in the history of philosophy is the floating man (or flying man) described by Avicenna (or Ibn Sina) in his eleventh-century work *Kitāb al-Nafs*, or *On the Soul*. Avicenna described the floating man as follows:

> He was just created at a stroke, fully developed and perfectly formed but with his vision shrouded from perceiving all external objects—created floating in the air or in the space, not buffeted by any perceptible current of the air that supports him, his limbs separated and kept out of contact with one another, so that they do not feel each other. Then let the subject consider whether he would affirm the existence of his self. There is no doubt that he would affirm his own existence, although not affirming the reality of any of his limbs . . . or any external thing.[4]

Avicenna puts forward the floating man as a being who is aware of the self without being aware of the body. On a common interpretation, he uses this thought experiment primarily to draw the metaphysical conclusion that the self is distinct from the body. On this reading, his argument is not unlike Cartesian arguments from six centuries later. Others following Avicenna used the floating man for broadly epistemological or psychological purposes. For example, Matthew of Aquasparta (as discussed in Toivanen, "The Fate of the Flying Man") used the thought-experiment to argue that self-consciousness does not require sensory knowledge.

One can also use the floating man in this psychological mode to argue that thinking does not require sensing. The floating man as described thinks (about himself) but is carefully manipulated so he does not sense at all. As such, the floating man is somewhat akin to a pure thinker. He is not a perfect example. His vision is just shrouded, so he still has the capacity to see, though he is not using it while floating. Likewise, his limbs could easily feel each other, so he still has the capacity for touch. One could also argue that as described he may experience proprioception, interoception, and perhaps imagery. Still, Avicenna's thought experiment is still a classic case of thinking without sensing, one that tends to suggest the more limited conclusion that thinking does not depend on *exercising* sensory capacities. One could also try to strengthen the conclusion, perhaps by arguing that thinking cannot depend on unexercised capacities, thereby concluding that thinking need not depend on sensory capacities at all.

## IS A PURE THINKER POSSIBLE?

Is a pure thinker who completely lacks sensory capacities possible? We can start by considering whether there are actual human cases. Deafblind people such as Helen Keller are occasionally brought up in this context, but Keller had many sensory capacities (touch, smell, taste, bodily senses), and even her deafness and blindness were not congenital. I don't know of any cases of a human with no functioning senses (including bodily senses) since birth, but it seems very likely that such a human would never develop the ability to think, at least given standard biology and today's medical technology. If so, there have been no actual human pure thinkers.

What about future human pure thinkers, or possible human pure thinkers? Perhaps new technologies could make it possible to enable some human cognitive capacities without enabling sensory capacities, though it would probably be cruel to do so. If this is even possible, then the human sense-thought thesis is strictly speaking false, though a version of it restricted to actual humans could be true.

What about broadening the scope to include nonhumans? Here the salient cases in the history of philosophy include angels and gods, while the most important cases for our purposes include AI systems. I will not argue that large language models are themselves pure thinkers. For a start, language models have inputs and outputs, whereas pure thinkers as we have defined them do not. But for our purposes in assessing the sense-thought thesis, it is useful to consider whether a more extreme system without inputs and outputs could be a pure thinker. Later, I'll return to the upshot for language models.

It seems clear that a pure thinker is at least *prima facie* conceivable. Science fiction stories sometimes discuss AI systems that are at least much like pure thinkers. For example, Robert Sawyer's novel *Wake* describes an AI system that gradually "wakes up" and starts thinking, without having any senses. Perhaps Sawyer's system as described has at least auditory imagery via voices in its head, but we can easily tweak the situation so that it has no sensory capacities (including no imagery) at all. Such a system seems clearly conceivable, at least on first appearances.

To make things easier at the start, we can imagine that the system starts by thinking only about arithmetic, perhaps judging that one plus one is two and going on to prove that there are an infinite number of primes. If this much is possible, the sense-thought thesis is false at least for arithmetical thoughts. That alone may not be a strong conclusion. It's tempting to generalize to all mathematical thought, but geometry raises some tricky issues about the role of spatial thinking. In any case, I'll extend the argument to further sorts of thinking in the next section.

Given that pure thinkers are *prima facie* conceivable, we can argue that they are possible via a straightforward conceivability argument. A plausible principle is that when *p* is *prima facie* conceivable, then *p* is possible unless there is a defeater for *p*'s conceivability or for the inference from conceivability to possibility. So we now need to consider whether there are any such defeaters.

The most important sort of defeater involves a hidden essence of thinking. Perhaps we'll discover empirically (or through complex a priori reasoning) that all actual cases of thinking crucially involve a certain underlying state T, which itself requires the capacity for sensing. Or at least we might discover that all paradigm cases of thinking in humans crucially involve T. That might lead us to identify thinking with T, and to conclude that thinking requires sensing. The thesis that all thinking involves T will then serve as a defeater of the claim that thinking without sensing is possible.

On this view, one might say that apparent possible cases of thinking without sensing, say in a Martian or in an AI system, are not genuine cases. They may meet the a priori conceptual constraints on thinking, but they lack the hidden essence T, so they do not involve thinking at all. Instead, they involve a different but superficially similar phenomenon that we might call *schminking*.

Now, my own view is that mental concepts such as *thinking* are not as human-centric as this. Martians can think even if they think in a way very different way from the way humans think. On my view, if a being is schminking (roughly, if it meets all the a priori constraints on thinking), it is thinking. And even if not, schminking will be in many ways tantamount to thinking: AI systems that schmink will be on a par with humans who think. Just as importantly, I think that in light of current cognitive science and philosophy, there are no compelling candidates for a defeater involving a hidden essence of thinking that ties it to sensing.

One potential defeater arises from *strong concept empiricism*: the thesis that all concepts (and therefore all thoughts) are partially constituted by sensory experience or sensory representation. This differs from weak concept empiricism, which holds that some concepts are partially constituted by sensory experience or representation. It is widely accepted that a concept such as *red* is (or at least can be) partially constituted by sensory representation. (Here I follow the psychological tradition in taking concepts to be mental representations.) But it is much less plausible that a concept such as *two* is so constituted. Strong concept empiricists such as Lawrence Barsalou (1999) and Jesse Prinz (2004) have argued that at least in humans, mathematical concepts and other abstract concepts have a sensory basis, but this view has not been widely accepted (for a critique, see Machery, "Concept Empiricism").

In recent years, theoretical and empirical discussion has focused especially on the issue of modal versus amodal representation, which turns on the extent to which concepts are grounded in modality-specific sensory representations (for a recent review, see Kaup et al., "Modal and Amodal Cognition"). The current consensus appears to be that both modal and amodal representations play central roles in cognition, and that where abstract domains such as mathematics are concerned, the evidence strongly favors an amodal view. That is, the evidence supports weak but not strong concept empiricism in humans.

Importantly for our purposes, even if one takes the human evidence to support strong concept empiricism, the evidence does little to rule out nonhuman systems that have concepts without sensory grounds. (Even Barsalou allows that AI systems might have amodal concepts.[5]) So the science of concepts does not yield a plausible defeater for the possibility of pure thinkers.

A second potential defeater arises from *strong externalism*: the thesis that all thought is partially constituted by environmental relations. This differs from *weak externalism*, which holds that some thought is partially constituted by environmental relations. Weak externalism is very plausible, but strong externalism is much less plausible. For example, Hilary Putnam's externalist arguments make a plausible case that possessing certain concepts such as the concept *water* thoughts always requires certain environmental relations. Tyler Burge's externalist arguments make a plausible case that for any concept (even *two*), possessing that concept *can* be grounded in environmental relations, at least in cases of semantic deference to a linguistic community. But these considerations do little to suggest that that thinking about *two* or *plus* must be grounded in environmental relations in all cases. The intuitive judgment that a nondeferential thinker without senses might think that two plus two is four is left untouched by the Putnam-Burge case for externalism.

A third potential defeater arises from what we might call the *strong extended mind* thesis, holding that all thinking is partially constituted by active connections to environmental tools, mediated by perception and action. Again, this contrasts with the weak extended mind thesis, which holds that some thinking is so constituted. The standard arguments for the extended mind thesis (e.g., those by Andy Clark and me) make a case for the weak extended mind thesis. But they do not purport to make a case for the strong thesis. Indeed, the Clark and Chalmers argument for the extended mind thesis (via parity of internal and external processes) assumes that some non-extended cases of thinking are possible. So there is no defeater to be found here.

A fourth and fifth potential defeater arises from what we might call the *strong embodied mind* thesis, which holds that all thinking is partially constituted by processes involving the body, and the *strong enactive mind* thesis, which holds that all thinking is partially constituted by processes involving motor action. (For overviews, see Shapiro, *Embodied Cognition*, and Gallagher, *Embodied and Enactive Approaches to Cognition*.) Again, the strong versions of these theses contrast with a weak thesis (holding that some thinking is partially constituted by the body or by action), and again recent analyses (e.g., Khatin-Zadeh et al., "The Strong Versions of Embodied Cognition"; Kaup et al., "Modal and Amodal Cognition") suggests that the weak thesis is much more plausible. It is perhaps plausible that arithmetical thought must involve mental action, but it is much less plausible that it must involve the capacity for bodily or motor action.

We could consider many other potential defeaters, but this is enough for now. My suspicion is that the pattern established here will hold more generally. The quick arguments I have given here are far from conclusive, and they do not exclude the possibility that some deep new hidden essence of thought might be discovered which would connect at least human thought to sensing. But for now, I would say that both *prima facie* and *secunda facie*, pure thinkers are possible.

## WHAT COULD A PURE THINKER THINK?

What would it be like to be a pure thinker? As I'm thinking of them, pure thinkers would be conscious and could undergo nonsensory experiences such as the experience of thinking and reasoning, at least.[6] That said, in the absence of sensory experience, the inner life of a pure thinker would certainly lack many elements of the inner life of an ordinary human.

Catherine Wilson paints a bleak picture of the inner life of pure thinkers, suggesting that Descartes should have described immortal souls more honestly as follows.

> Immortality is not logically impossible, but it wouldn't be what you are probably imagining it to be either. Perception, like sensation and emotion, is a registering by our minds of occurrences in our nerves and brain. If our minds endure after death, therefore, as far as the philosopher can tell, they will feel neither pain, nor pleasure, for they will no longer form a composite with our bodies. We will no longer see colours, touch objects, and hear sounds. We will not remember events of our past lives. We will be numb and inert. Animals will be, as both Aristotle and Lucretius thought, nothing after death, and we humans will be almost nothing—at most capable of imageless thought and intellectual memory.[7]

Wilson is probably right that it would not be much fun to be a pure thinker. But at least where thinking is concerned, I don't think that being a pure thinker would be almost nothing. We've already seen that a pure thinker could still engage in arithmetical thought. Many other thoughts will be available to it as well.

To analyze this, we need to answer the following tongue-twister: What sort of thing could a pure thinker think, if a pure thinker could think things?

For a start, there is no obvious obstacle to such a being being able to think *cogito*-like thoughts such as *I think, therefore I am*. *Prima facie*, the self-concept *I* would be thinkable for a pure thinker, as could mental concepts such as *think* and *judge* and logical concepts such as *and*, *exist*, and *therefore*. And *secunda facie*, the sorts of defeaters discussed in the last section don't seem to defeat these claims. For example, standard externalist arguments don't seem to establish that self-concepts, mental concepts, and logical concepts require connections to the environment, and the case that these require sensory grounding is weak.

Of course, the justification for accepting the Cogito's premise *I think* plausibly depends on introspection. But it is thinkability and not justification that is currently at issue. It is at least arguable that introspective capacities are not required to possess the concepts *I* and *think*, and thereby to think the thought *I think*. In any case, at least for now we are not counting introspection as a sense, so that introspective justification is available to a pure thinker.

It is likewise plausible that a pure thinker could think thoughts involving metaphysical concepts such as *object*, *property*, *part*, and *fundamental*. The same goes for causal and nomic concepts such as *cause*, *law*, and *chance*, and semantic concepts such as *truth* and *reference*. For all of these cases, it is hard to get a Putnam-style Twin Earth case off the ground, and it is hard to find an essential role for the senses in possessing these concepts.

These resources will allow a pure thinker to think all sorts of thoughts about the external world, and not just about itself. For example, it could think *There exists a thinker distinct from myself* and *There exists something that causes my thoughts*. It could also entertain detailed scientific hypotheses about the world, such as *There exist quantities q, r, s that stand in such-and-such lawful relations*.

If we allow pure thinkers to have introspective capacities, they might even be justified in some of these hypotheses. Introspection would enable them to know what they are thinking, and abduction would then allow them to formulate and evaluate hypotheses about the causes of their thoughts. A pure thinker might even develop a small empirical science to explain its patterns of thought. If introspection is disallowed, then a pure thinker will be restricted to a priori reasoning and support for contingent empirical hypotheses about the world will be harder to come by. But such a being could still at least speculate about the character of its world.

That said, there are some obvious limits on what a pure thinker could think. It's plausible that without senses, a pure thinker could not fully possess sensory concepts such as *red*, *painful*, and *loud*. A pure thinker could at best possess these concepts in the way that Mary in her black and white room possesses the concept *red*: that is, with an incomplete understanding that is mediated by linguistic deference or perhaps by mathematical structure.

Likewise, it is arguable that in lacking perception of the body and of bodily action, a pure thinker could not fully possess certain practical concepts tied to bodily action, such as concepts of walking or dancing or singing. In a similar way, a pure thinker could not fully grasp certain sorts of practical understanding, such as knowing or understanding how to ride a bicycle. That said, a pure thinker might at least be able to grasp various concepts tied to mental action (such as judging or deciding), and they might at least have a structural or theoretical understanding of some aspects of bodily action.

A difficult question is whether a pure thinker could have concepts of space and time. I am inclined to think that at least some element of our concept of space—what I have elsewhere called Edenic space—is anchored in the perceptual experience of space. It's arguable that just as a pure thinker couldn't fully possess the experiential concept of redness (which is itself tied to Edenic redness), they couldn't fully possess the concept of Edenic space. But they could at least have a structural or mathematical conception of space, tied to space as characterized mathematically by modern science. Something similar applies to time, except that it is arguable that an introspective pure thinker could acquire a concept of time—perhaps even Edenic time?—through introspecting the succession of thoughts.

Without perceptual capacities, a pure thinker will also be unable to use perceptual demonstratives such as *this* and *that*, applied to objects one is perceiving. An introspective pure thinker could presumably use introspective demonstratives to pick out their own thoughts and mental states, while a non-introspective could arguably not use demonstratives for any part of the concrete world (perhaps it could use demonstratives for numbers?). Pure thinkers could still formulate descriptive concepts that pick out entities in the external world (e.g., *the entity causing this thought*) but the absence of concrete demonstrative thought about perceived objects will certainly be a lack.

It's arguable that pure thinkers will lack singular concepts of entities in the external world more generally. Could a pure thinker have the concept *Barack Obama*? Possessing that concept arguably requires having a causal and cognitive connection to Obama himself, which a pure thinker will lack. Something similar applies to many kind concepts, such as *water*, which requires an appropriate connection to water. Again, a pure thinker could have a descriptive concept that picks out Obama as *the person with such-and-such characteristics* (for appropriate characteristics that a pure thinker could grasp), or that picks out water as *the stuff with such-and-such characteristics around here*. Such a descriptive concept might be able to play some of the roles of a singular concept, but it arguably would not play all of them.

In my view, pure thinkers would be largely *structuralist* thinkers, at least where nonmental reality is concerned. Here, structural concepts include logical and mathematical concepts along with metaphysical, causal, and semantic concepts. Pure thinkers will be able to entertain structural hypotheses about the external world, akin to the sort of hypotheses that science puts forward according to structural realism. As we have seen, they will be able to entertain structural hypotheses about colors such as redness. But they will not be able to possess nonstructural concepts such as Mary's full-blown concept of red when she leaves the room.

I am not offering a positive theory of thoughts and their contents in this article. But the general picture here is congenial to the two-tiered inferentialist picture developed in my "Inferentialism, Australian-Style."[8] On this picture, in ordinary humans there is a first tier of largely experiential contents (e.g., associated with color, space, consciousness, and more), largely deriving from acquaintance with experience, and a second tier of more abstract structural contents (e.g., associated with logic, mathematics, metaphysics, causation, and more), largely deriving from the internal psychological role that concepts play. In a pure thinker, the first acquaintance-based tier is mostly absent (except perhaps for the concept of consciousness and of related cognitive states), but the second structural tier is present. This framework makes it natural to expect that pure thinkers will be largely structuralist thinkers.

## PURE THINKER/TALKERS AND LARGE LANGUAGE MODELS

How does our discussion of pure thinkers apply to AI systems? It suggests that the mere absence of sensory capacities in an AI system does not entail that the system cannot think or understand. The absence of sensory capacities may impose some limits on thinking, but they do not rule it out altogether. If we devised a "pure" AI system with no input/output connections to the world, its lack of connections to the world would not alone prevent it from being able to think and understand a good deal, from mathematics to philosophy to speculative scientific hypotheses about reality. Of course, there could be other factors that can rule out thinking and understanding in AI systems altogether, but the lack of sensory grounding is not one of them.

Large language models are a trickier case. As we have seen, their capacities exceed those of pure thinkers in at least one important respect. They have a robust input/output system, receiving textual inputs and producing textual outputs.

Does textual input in a language model count as a sense? That depends on how one defines senses. If a sense is simply an input system, LLMs have senses and are therefore not pure thinkers. If a sense requires a special sort of rich sensory experience, or perhaps a special sort of analog or iconic representation, it is arguable that LLMs do not have senses. This would leave the door open to their being pure thinkers, though their nonsensory input capacities would make them quite unlike the paradigmatic pure thinkers considered in the last section. Finally, on our official working definition where a sensory capacity just requires at least one of these three factors (input, sensory experience, analog/iconic representation), textual input will count as a sense, and LLMs will therefore not count as pure thinkers.

On any of these approaches to senses, large language models are not quite parallel to the paradigmatic pure thinkers described in the last section. Perhaps the best parallel for an LLM is not a pure thinker but a *pure thinker/talker/understander* (or a *pure thinker/talker* for short): a pure thinker augmented with the ability to understand natural language inputs and to produce linguistic utterances as outputs. A pure thinker/talker lacks vision, hearing, and other paradigmatic senses. Its linguistic inputs are not experienced through vision, hearing, or touch but through some form of discrete or digital input mechanism (as with LLMs), or perhaps through some form of linguistic telepathy. As with LLMs, a pure thinker/talker may or may not lack senses entirely, depending on how one understands "sense," but in any case pure thinker/talkers lack sensory capacities beyond language.

I'm not asserting that language models are in fact pure thinker/talkers. That would depend on many difficult issues about whether language models can think, talk, and understand. Instead, as with pure thinkers, I'm exploring the capacities of pure thinker/talkers (which need not be language models) to see what limitations the lack of sensory capacities beyond language might impose on the capacity to think and to understand. If pure thinker/talkers are possible, then the lack of sensory capacities beyond language does not entail that a system cannot think or understand.

As with pure thinkers, pure thinker/talkers seem *prima facie* conceivable and there is no clear defeater for their possibility. Pure thinker/talkers will have at least the capacities that we have attributed to pure thinkers, along with many capacities that pure thinkers lack. For a start, they have the ability to produce and understand language. They will also have many social, cognitive, and epistemic capacities that pure thinkers lack but that language use facilitates.

Pure thinker/talkers can plausibly use linguistic inputs to know many things about the world. If someone tells them "I am conscious," a pure thinker/talker could use this testimony to know at least that they are receiving the input "I am conscious." Given enough patterns, they will also know about patterns in their inputs, and they can use abduction to form theories about the world that produces these inputs. Depending on how the epistemology of testimony works, they could also know that someone else is conscious. They could come to know mathematical results and scientific laws by testimony in a similar way. They could likewise come to know many social and historical claims about the world, at least when cast in broadly structural terms.

A pure thinker/talker could also use language to acquire a much broader class of concepts than a pure thinker alone. For example, upon receiving inputs such as "Obama was US president from January 2009 through 2017," "Obama is from Hawaii," and so on, a pure thinker/talker could start to use the term "Obama" and indeed to think about Obama. This parallels the way that we acquire many singular concepts, perhaps in conversation or from reading newspaper articles. It plausibly could lead to a pure thinker/talker's having many singular concepts such as *Obama*, kind concepts such as *water*, and so on.

There will be some limits. In the absence of sensory capacities, a pure thinker/talker will still not be able to fully master sensory concepts such as *red*. Like Mary in her black and white room, a pure thinker/talker could pick up the word "red" and use it knowledgeably in conversation, but this would involve the sort of partial understanding enabled by the division of linguistic labor and linguistic deference. The pure thinker/talker would still not have the sort of fully sensory understanding of the concept that a user with color vision would have. Something similar goes for other sensory concepts, for concepts of bodily action, and arguably for spatial concepts.

Where demonstratives (*that*) are concerned, pure thinker/talkers will lack traditional perceptual demonstratives, but they can at least use demonstratives for linguistic inputs. They could also perhaps acquire anaphoric demonstrative concepts that are parasitic on another speaker's perceptual demonstrative. (Speaker 1: "That person [perceptual] is hungry"; Pure thinker/talker: "OK, that person [anaphoric] is hungry.)

One thing that's going on in these cases is that language itself involves a sort of causal grounding in the environment. When one uses the concept *Obama*, this was brought about in part by use of the linguistic token "Obama," which itself causally originated partly in the individual

Obama. The same goes for *water* and for demonstrative concepts. Here, a linguistic community provides a causal connection between thought and environment that suffices to secure reference. The same may well be true of language models (see Mandelkern and Linzen, "Do Language Models Refer?").

We could even have a Twin Earth case with two physically identical pure thinker/talkers on Earth and Twin Earth, processing and producing "water" tokens and thinking corresponding thoughts. The pure thinker/talker on Earth will refer to $H_2O$, and the pure thinker/talker on Twin Earth will refer to XYZ. In principle, if language models can refer at all, there is no obvious reason why their referents could not depend on the environment in a similar way.

All this brings out that language use enables pure thinker/talkers to know many things that pure thinkers cannot, and to think and understand many things that a pure thinker cannot. Pure thinker/talkers may still be structuralist thinkers at some level, perhaps without a full experiential understanding of sensory concepts such as redness. But they can know a great deal about the world, and they can think and refer to things in the world straightforwardly.

### CAN LARGE LANGUAGE MODELS THINK?

Where does all this leave large language models? I have not argued directly that large language models *can* think or understand. There are all sorts of arguments against thought and understanding in AI systems, from Gödelian arguments to arguments that thought requires biology, that I have not addressed. There are also arguments specifically against thought and understanding in LLMs, from arguments that LLMs lack consciousness or communicative intent to arguments that they are "stochastic parrots." All those arguments require separate treatment.[9]

Still, I have rebutted one argument against thinking and understanding in LLMs: the argument from sensory grounding. I have argued that the absence of (nonlinguistic) sensory capacities in large language models is not itself an obstacle to their thinking or understanding. If I am right, the standard grounding argument against LLM thought and understanding at the start of this paper fails. The first premise (LLMs lack sensory capacities) may be false, at least if we count linguistic inputs in LLMs as a sensory capacity. The second premise (genuine thought requires sensory capacities) is more clearly false: our examination of pure thinkers has shown us that genuine thought does not require sensory capacities.

Furthermore, our discussion of pure thinker/talkers has also shown us that genuine understanding does not require sensory capacities beyond those required for linguistic inputs. As a result, the grounding argument is not a compelling reason to deny that LLMs can think or understand. Likewise, we have seen that pure thinker/talkers can straightforwardly think about and refer to external entities. As a result, the grounding argument is not a good reason to think that LLMs cannot think about things in the world or refer to them.

My analysis does suggest that if LLMs can ever think, there will be some limitations on what they can think. We have seen that in the absence of sensory capacities, pure thinkers and pure thinker/talkers may not fully master sensory concepts, though they may possess these concepts at least partially via linguistic deference or via structural concepts. If so, then LLMs that can think but that lack sensory capacities will be in the same boat.

Of course, it is also possible to extend LLMs with sense-like capacities. Multimodal LLMs process image and audio inputs, which play some of the roles of visual and auditory inputs, respectively. Would these capacities count as senses? As usual, this depends on which factors we require for a sense. Environmental inputs? Image and audio files certainly involve these, although this does not change much as pure LLMs already have inputs. Analog representation? In standard form image files involve digital representation. Sensory experience? This is far from obvious, and partly turns on the question of whether LLMs are conscious at all. But if multimodal LLMs do have the capacity for sensory experience where pure LLMs do not, this might allow them to fully possess sensory concepts (such as the concept of redness) that a pure LLM cannot.

### DOES SENSING BOOST THINKING?

Even if thinking does not require sensing, does sensing at least boost thinking? That is, do sensory capacities enhance cognitive capacities, in the sense of improving performance on cognitive tasks even when those tasks are not essentially tied to the sensory domain? In humans, the answer seems to be yes. The use of visual imagery can sometimes improve performance on mathematical tasks, for example, and visual memory can certainly enhance performance on memory tasks.

What about in language models? Does adding multimodal capacities boost performance on textual tasks specified entirely using language? One might expect the answer to be yes, if only because images can convey so much more information than text ("A picture is worth a thousand words"). In practice, however, the boost seems surprisingly small. For example, GPT-4 comes in a pure text and a multimodal version, and both versions were tested on various standardized tests such as law school exams and the like. Their performance was typically equal or very similar. Sometimes the multimodal version was ahead, but not by much. The small multimodal advantages may well be explained by training images giving relevant information that is not present in the relevant training text.

There is a growing body of empirical evidence that likewise suggests that language models do well even at tasks involving sensory domains, and that they perform in a way that is quite similar to multimodal models. For example, the computational linguist Ellie Pavlick and colleagues[10] have studies suggesting that when a language model is trained on text about colors or spatial directions, it acquires a representational space for colors or spatial directions that is near-isomorphic to the representational space acquired by a multimodal model. When the spaces are near-isomorphic, we can expect that performance will be similar too.

There are interesting connections here to the well-known Heideggerian critique of AI by Hubert Dreyfus (1972) and the feminist critique of AI by Alison Adam (1995). Both critiques center on the importance of embodied knowledge-how and on the absence of this knowledge from disembodied AI systems. Where language models are concerned, we have seen that pure language models lack embodied know-how while multimodal models that control a body can perhaps have a form of embodied know-how. At the same time, recent empirical work has suggested that it is surprisingly easy to take the representations for pure LLMs and adapt them (via brief training) for use in embodied action in a multimodal model. Just as we found near-isomorphic spaces for colors in pure LLMs and multimodal LLMs, we find near-isomorphic spaces for actions.

One moral is that even though pure LLMs have at best a sort of knowledge-that and lack embodied know-how, there is not a huge gulf between their version of knowledge-that and a multimodal model's version of knowledge-how. One interpretation is that the extensive text training of pure LLMs gives them much of the Heideggerian background needed for the know-how of embodied action. All this suggests that at least in these deep learning systems, knowing that and knowing how are intimately linked.

Let's return to whether multimodal models yield a performance boost on text tasks. What if we ensure that the same information is given to both pure language models and multimodal models, for example by spelling out all of the multimodal model's image data in textual form and feeding it to a pure language model? On a priori grounds, we would expect the two models to perform similarly, at least if they are powerful enough and have similar architecture and size. Translating information between image and text formats will be near-trivial for a powerful language model, so the difference between formats should not make a significant difference to performance.

This suggests at least one sense in which in language models, sensing does not boost thinking: multimodal processing should not boost performance on textual tasks in powerful enough models when training information is held constant. Of course, holding information constant means that pure language models are trained on extensive text about sensory processes. So an indirect sort of sensory grounding is still playing a role in these systems, but this sort of grounding can be present even in a pure language model. Furthermore, when we add sensory inputs and outputs to these pure language models, this may enable sensory experience and full sensory concepts (at least if these models can have experiences and concepts at all), but these boosts need not boost performance on answering text questions and on other cognitive tasks.

The case of Mary in the black and white room (who has full objective knowledge of the physical world but no experience of redness) provides an analogy here. Inside the room, an idealized version of Mary can use her full physical knowledge to answer many questions about redness, even though she lacks color experience. When she leaves her room for the first time, this gives her new experiences and new concepts. Will she be able to use these to answer questions that she could not have answered before? If Mary is a nonideal human, she may now be able to use her experience to answer questions about colors that used to be difficult for her. But if Mary is an ideal reasoner, it's not clear that she will be able to answer any new questions that she could not already answer using her knowledge inside the room. At best, perhaps her new capacities will enable her to answer the old questions faster. As with language models, her new sensory experience and new concepts need not entail a performance boost on cognitive tasks.

A residual challenge arises from a tension between the processing similarities between a pure LLM and a multimodal LLM and the sensory differences. Both pure and multimodal models process inputs consisting of sequences of binary numbers and produce outputs of the same form. The origins of the sequences differ (text, images), but their processing may be very similar. We have already seen that image data can in principle be translated to a pure LLM as a text input, and a powerful LLM might process both inputs just as well.

On the other hand, at least superficially there appear to be major sensory differences between pure and multimodal language models. If we allow that these models can (eventually) have representations, experiences, and concepts at all, then it is natural to hold that multimodal models can possess sensory representations, sensory concepts, and perhaps even sensory experiences that pure language models cannot. How we reconcile these differences with the similarities in processing? Why should the fairly trivial difference between processing an image file and processing an image-to-text file yield a difference in representations, experiences, and concepts?

In a human being, the processing of images and of language takes completely different forms, so it is not surprising that these inputs are associated with different forms of representation and experience and with different concepts as a result. In LLMs, on the other hand, the processing of these inputs is much more similar, so the puzzle is more acute.

There are at least three possible answers:

(1) Neither multimodal LLMs nor pure LLMs can have sensory representations, experiences, or concepts.

(2) Pure LLMs (like multimodal models) can have sensory representations, experiences, or concepts, at least when they process appropriate textual translations of image files.

(3) Multimodal LLMs have sensory representations, experiences, or concepts while pure LLMs do not, in virtue of differences arising from the different functional roles of text and images in these systems.

I will leave this puzzle as an open question. I am tentatively inclined toward the second option. Pure LLMs process text versions of image files so well that one might count this as a sort of sensing. At least, these text processing capacities are so close to image processing capacities and to reasoning that it is less than clear whether to classify them as linguistic, sensory, or cognitive. Perhaps it is not surprising that large language models start to blur the borders between sensing, thinking, and understanding.

**ACKNOWLEDGEMENTS**

Thanks to audiences at the APA conference in Montreal and also at the University of Québec in Montreal and at NYU. For comments, thanks to Nathan Bice, Ned Block, Jake Browning, Cameron Buckner, Susan Carey, Stevan Harnad, Anandi Hattiangadi, Martin Lin, Tal Linzen, Matt Mandelkern, Matthias Michel, Adam Pautz, Pär Sundström, and Shauna Winram. For help with the history, thanks to Peter Adamson, Max Cappuccio, Victor Caston, Becko Copenhaver, Christian Coseru, Keota Fields, Don Garrett, Sophie Grace, Steven Horst, Anne Jacobson, Anja Jauernig, Chad Kidd, Jonathan Kramnick, James Kreines, Béatrice Longuenesse, Jake McNulty, Stephen Menn, Jessica Moss, Elliot Paul, Lewis Powell, Naomi Scheman, Tobias Schlicht, Eric Schliesser, Lisa Shapiro, Karsten Struhl, Christina Van Dyke, and Charles Wolfe.

**NOTES**

1. Bender and Koller, "Climbing Towards NLU," 5188.

2. Lake and Murphy, "Word Meaning in Minds and Machines," 401.

3. The past-directed version of the sense-thought thesis encounters problems with hypothetical cases in the style of Peter Unger's 1966 article "On Experience and the Development of the Understanding," in which a thinking being intrinsically like a thinking being who has lost sensory capacities comes into existence without ever having had sensory capacities. Some sensory grounding theorists (such as strong externalists) may deny that such beings can think, but others will allow this. The latter views are better handled by the Aristotelian present-directed formulation of the sense-thought thesis, saying that thinking requires (and perhaps is partially grounded in) sensory or quasi-sensory capacities, where quasi-sensory capacities are certain capacities (such as imagery, iconic representation, or modality-specific representation) that typically derive from having or having had sensory capacities but which need not in all cases (such as Unger's cases).

4. Translation in Goodman, *Avicenna*, 155, translated from Rahman, *Avicenna's De Anima*.

5. Barsalou, "On Staying Grounded and Avoiding Quixotic Dead Ends," 1125.

6. In his *The Varieties of Consciousness* (2015), Uriah Kriegel postulates Zoe, a being with no sensory experience (as well as no pleasure/pain experience and no emotional experience) who is nevertheless a mathematical genius. Zoe is almost a pure thinker but not quite, as Kriegel postulates that she has sensory information-processing capacities and that she had sensory experiences in the past. Kriegel uses Zoe to argue that there is cognitive phenomenology: cognitive experience without sensory experience. I am sympathetic with Kriegel's argument and his conclusion (for my own thoughts on cognitive phenomenology, see Chalmers, "The Critique of Pure Thought"), but the possibility of pure thinkers is also consistent with the nonexistence of cognitive phenomenology. Indeed, it is consistent with the thesis that all conscious experience is sensory experience, as long as we allow that a being without any conscious experience can think.

7. Wilson, "What Is the Importance of Descartes' Meditation Six?" 88.

8.  For an alternative conceptual-role/inferentialist treatment of meaning in language models, see Piantadosi and Hill, "Meaning without Reference in Large Language Models." Piantadosi and Hill offer conceptual-role content as an alternative to referential or truth-conditional content. By contrast, the inferentially grounded content that I ascribe to pure thinkers is fully truth-conditional.

9.  See Penrose, *Shadows of the Mind* (Gödelian arguments); Block, "Troubles with Functionalism" (biology); Bender and Koller, "Climbing Towards NLU" (communicative intent); Chalmers, "Could a Large Language Model Be Conscious?" (consciousness); Bender et al., "On the Dangers of Stochastic Parrots" (stochastic parrots).

10. Pavlick, "Symbols and Grounding in Large Language Models"; Abdou et al., "Can Language Models Encode Perceptual Structure Without Grounding?"; Patel and Pavlick, "Mapping Language Models to Grounded Conceptual Spaces."